\documentclass[runningheads]{llncs}
\usepackage{graphicx,subfigure,algorithm,algorithmic,amssymb,amsmath}
\begin{document}
\title{Stacking Ensemble Learning in Deep Domain Adaptation for Ophthalmic Image Classification}
%
%

\author{YEGANEH MADADI\inst{1,2}
\and
VAHID SEYDI\inst{3}
\and
JIAN SUN\inst{2}
\and
EDWARD CHAUM\inst{4}
\and
SIAMAK YOUSEFI\inst{2}
}

\authorrunning{Y. Madadi et al.}
%

\institute{ University of Tehran, Iran
\and
University of Tennessee Health Science Center, USA
\and
Bangor University, UK
\and
Vanderbilt Eye Institute, USA}


%
\maketitle              
\begin{abstract}
Domain adaptation is an attractive approach given the availability of a large amount of labeled data with similar properties but different domains. It is effective in image classification tasks where obtaining sufficient label data is challenging. We propose a novel method, named SELDA, for stacking ensemble learning via extending three domain adaptation methods for effectively solving real-world problems. The major assumption is that when base domain adaptation models are combined, we can obtain a more accurate and robust model by exploiting the ability of each of the base models. We extend Maximum Mean Discrepancy (MMD), Low-rank coding, and Correlation Alignment (CORAL) to compute the adaptation loss in three base models. Also, we utilize a two-fully connected layer network as a meta-model to stack the output predictions of these three well-performing domain adaptation models to obtain high accuracy in ophthalmic image classification tasks. The experimental results using Age-Related Eye Disease Study (AREDS) benchmark ophthalmic dataset demonstrate the effectiveness of the proposed model.

\end{abstract}
\section{Introduction}

In real-world applications, it is typically challenging to obtain sufficient number of annotated training samples. To address this problem, domain adaptation (DA) \cite{Madadi2020} has been successfully developed to adapt the feature representations learned in the source domain with required label information to the target domain with fewer or even no label information.

There are two main categories for deep domain adaptation approaches: Domain-invariant features adaptation, and discriminators adaptation. The first tries to map source and target domains in the common subspace to learn the shared features space approach by adding adaptation layers into deep neural networks \cite{zhu2019aligning,zhu2019multi}. The second approach attempts to adversarially recognize features in the variant domains by adding the domain discriminator \cite{li2021feature}.

Our proposed method is based on domain-invariant features adaptation. This category of methods is obtained through optimizing several measures
of domain discrepancy, such as Maximum Mean Discrepancy (MMD) \cite{zhu2019multi,kang2019contrastive,deng2020rethinking}, Low-rank representation \cite{ding2018deep,madadi2020deep}, and Correlation Alignment (CORAL) \cite{chen2019joint,rahman2020minimum,cheng2021robust}.
Furthermore, we propose a combination of deep DA methods through the stacking ensemble strategy. Stacking ensemble methods are an outstanding strategy in machine learning (win most Kaggle competitions) \cite{pavlyshenko2018using}, and extending them with domain adaptation models makes the technique very useful for solving real-world problems. Ensemble learning methods integrate multiple machine learning models (base learners) that each model is trained to solve the similar problems and then the outcome of base models are combined for achieving better results. As the outcome is the majority voting (in the case of classification), the models could be both more accurate and more robust.

\begin{figure}[!t]
\centering
\includegraphics[scale=.3]{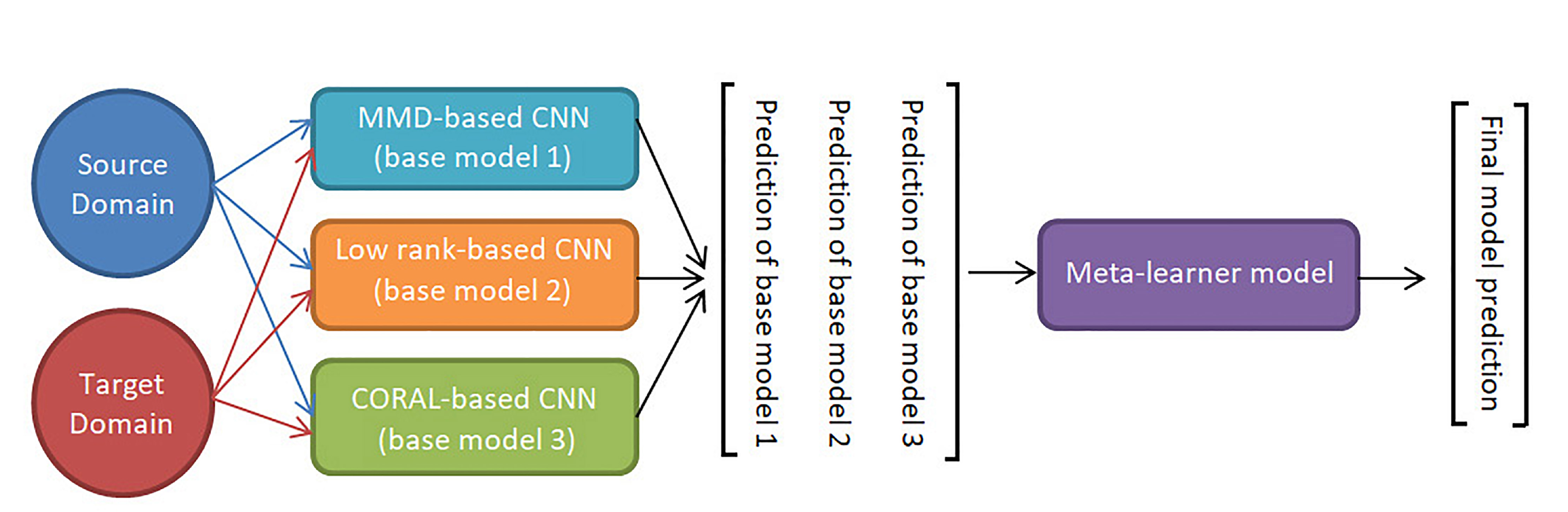}
\caption{Overview of the proposed model. Stacking combines multiple predictive models to generate a new combined model.}
\label{fig:1}
\end{figure}

We proposed a novel model, \textbf{S}tacking \textbf{E}nsemble \textbf{L}earning in \textbf{D}omain \textbf{A}daptation (SELDA), by introducing a deep domain adaptation method to acquire a cross-domain high-level feature representation and to reduce the cross-domain generalization error by the stacking learning method. In particular, we focus on the ophthalmic image classification task in an unsupervised scenario.
Our model includes three base DA models and a meta-learner model.
The model architecture for each of the three base DA models consists of domain-general and domain-specific representations across domains for unsupervised domain adaptation. For domain-specific parts, we apply a hybrid neural structure to extract multiple representations and extract more information from input images. Furthermore, to compute the adaptation loss and to decrease discrepancy between source and target domain distributions, MMD, Low-rank, and CORAL, are extended in base models. Our stacking DA model is illustrated in Figure \ref{fig:1}.

Our approach can be implemented via the most feed-forward methods and trained by using standard backpropagation.
The contributions of this paper are summarized as follows:
\begin{itemize}
\item{To the best of our knowledge, the proposed method (SELDA) is the first stacking model for deep domain adaptation in the ophthalmic image classification tasks.}
\item{We propose the multi-representation deep domain adaptation networks as base models that are ensembled through a stacking strategy to reach high accuracy.}
\item{The MMD, Low-rank, and CORAL are jointly extended to align the domain discrepancy in deep neural networks.}
\item{Extensive experiments demonstrate that SELDA achieves state-of-the-art performance on Age-Related Eye Disease Study (AREDS) \cite{study1999age} benchmark ophthalmic dataset.}
\end{itemize}

\section{Proposed Model}
\label{Proposed Model}
Domain adaptation is effective in situations that efficient labeled data in the target domain does not exist or is scarce. We propose a DA framework to improve the accuracy of classification tasks using an innovative stacking ensemble learning approach on ophthalmic datasets.

We are given a source domain $D_{s} = \{(\textbf{x}_{i}^{s},\textbf{y}_{i}^{s}) \forall i \in [1:n_{s}]\}$ where $(\textbf{x}_{i}^{s},\textbf{y}_{i}^{s})$ is tuple of source data and their labels, and $n_{s}$ is the number of labeled source samples, and a target domain $D_{t} = \{(\textbf{x}_{j}^{t}) \forall j \in [1:n_{t}]\}$ where $\textbf{x}_{j}^{t}$ is the target data, and $n_{t}$ is the number of unlabeled target samples. The source and target domains have different probability distributions. The purpose is to align these distributions by designing deep DA models.

Almost all DA models apply the single-representation structure, which focuses on the partial information from the data, but multi-representation structures can extract more information on the data. So, we learn multiple domain-invariant representations to obtain better performance where a hybrid structure with multiple substructures is utilized to extract multiple representations from input images.

Furthermore, we apply MMD, Low-rank, and CORAL techniques to reduce the distributions discrepancy between the multiple representations extracted from the source and target domains on three CNN models. We obtain higher accuracy by proposing a stacking ensemble learning approach on them.

We introduce these MMD-based, Low-rank-based, and CORAL-based deep DA models as base learners and learn these models on the training data. For each of the three base learners, predictions are made for observations on the validation data. Then, we propose a meta-learner model and fit it on predictions that are made by the base learners as inputs. Finally, we test the meta-learner model on testing data.

\subsection{Base Models}
\begin{figure}[!t]
\centering
\includegraphics[scale=.42]{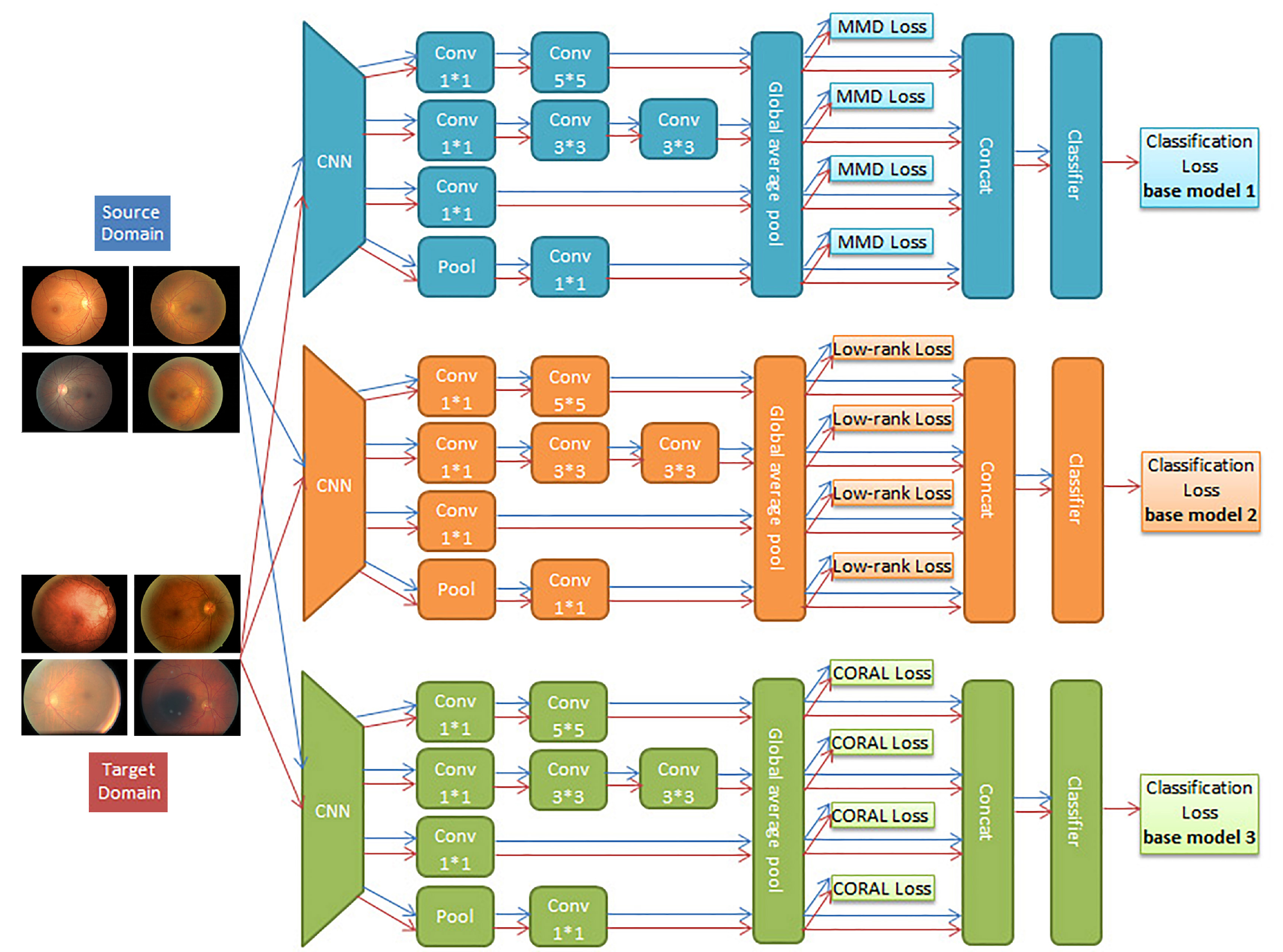}
\caption{The architecture of the base models in SELDA approach.}
\label{fig:2}
\end{figure}

The structures of three base models are similar, but the domain adaptation methods used to train the parameters are different. The architecture of each base model consists of three parts. The first part of each base model is the CNN, which is used to convert high-pixel images to low-pixel ones. The second part is the global average pooling for extracting representations from low-pixel images. Finally, the third part is the model prediction.
The architecture of base models is illustrated in Figure \ref{fig:2}. We have four types of convolution-pooling layers to extract different representations of the data. In each base model, one of the DA methods is applied to all different representations.

The optimization problem of domain adaptation is weighted sum of two cost functions. The first cost is used to minimize the classification errors on the source set, and the second cost minimizes the discrepancy between the source and target data in each base model.
Let $\textbf{X}^s$ be a matrix containing all training data of the source domain, wherein $i$th row corresponds to $i$th datum $\textbf{x}^{s}_i$. Similarly, $\textbf{X}^t$ is a matrix containing all target domain data. Furthermore, assume $g$ to be the general feature extractor, and $\{s_i\}_{i=1}^d$ be $d$ different specific feature extractors. Then, the cost function can be defined as
\begin{equation} \label{Equation:1}
  \noindent\footnotesize
  \min_{f,g,\{s_i\}_{i=1}^d} \dfrac{1}{n_{s}} \sum_{i=1}^{n_{s}} J\bigl(f([s_{1}(g(\textbf{x}^{s}_i));...;s_{n}(g(\textbf{x}^{s}_i))]), y^{s}_i\bigr)+ \lambda \sum_{i=1}^{n} D(s_{i}(g(\textbf{X}^{s})),s_{i}(g(\textbf{X}^{t}))),
\end{equation}
where $[s_{1};...;s_{n}]$ is the concatenated vector of different features, $f$ is a function from stacked features to scores of different labels, $J$ is the classification cost measuring the distance between label scores and true labels $y$, and $D$ is the cost  for minimizing the discrepancy between the source and target distributions. In this equation, $\lambda > 0$ indicates the trade-off parameter. In this paper, $f$ is a fully connected network followed by a softmax layer, and $J(.,.)$ denotes the cross-entropy loss.

The domain-general representation is implemented based on ResNet50. The domain-special representations for each base model are extracted by the substructure1 (conv1 $\times$ 1, conv5 $\times$ 5), substructure2 (conv1 $\times$ 1, conv3 $\times$ 3, conv3 $\times$ 3), substructure3 (conv1 $\times$ 1), and substructure4 (pool, conv1 $\times$ 1).
Since training deep CNN needs a large amount of labeled data that is expensive for many DA applications, so we utilize the CNN pre-training networks on ImageNet2012 data and then fine-tune them similar Long et al. \cite{long2017deep}.
The loss functions for minimizing the discrepancy between source and target domains can be MMD-based, Low-rank-based, and CORAL-based. These adaptation methods and training the parameters of each method are obtained as the following subsections.

The model training applies standard mini-batch stochastic gradient descent (SGD) method. In each mini-batch, the equal number of source domain data and target domain data are sampled to solve the bias which is caused by domain size.

\subsubsection{Maximum Mean Discrepancy (MMD):}
\label{Maximum mean discrepancy}
MMD is a metric widely used to measure the discrepancy of marginal distributions. By minimizing the MMD metric in the following equation, the marginal distributions between the source and target domains become close:
\begin{equation} \label{Equation:4}
  \noindent\footnotesize
  D_{MMD}(\bar{\textbf{X}}^{s},\bar{\textbf{X}}^{t}) = \biggl \| \frac{1}{n_{s}} \sum _{x_{i} \in \bar{\textbf{X}}^{s}} \Phi(\textbf{X}_{i}) - \frac{1}{n_{t}} \sum _{x_{j} \in \bar{\textbf{X}}^{t}} \Phi(\textbf{X}_{j})\biggl \|_{H}^{2},
\end{equation}
where $\Phi$ represents the kernel function, and $||.||_H$ is the norm in the Hilbert space.

Minimizing the difference between the conditional distributions of source and target domains is definitive for robust distribution adaptation. So we utilize conditional MMD (CMMD) instead of MMD to decrease domain discrepancy. We apply CMMD to the first base model for measuring the domains discrepancy $D$ in Equation (\ref{Equation:1}) identical to \cite{zhu2019multi}. Here we calculate the distance among the class conditional distributions $P(x_{s}|y_{s} = c)$ and $Q(x_{t} |y_{t} = c)$, which is called CMMD. Each class label in the source domain and each pseudo class label in the target domain is represented by $c \in \{1,...,C\}$. The output of the deep NN, $\hat{y}_{i}^t = f(\textbf{X}_{i}^t)$, could be utilized as the pseudo label for target data. We expect to iteratively improve the quality of pseudo labels of the target domain during the optimization.
\begin{equation} \label{Equation:4-1}
  \noindent\footnotesize
  D_{CMMD}(\bar{\textbf{X}}^{s},\bar{\textbf{X}}^{t}) = \frac{1}{C} \sum _{c=1}^{C} \biggl \| \frac{1}{n_{s}^{(c)}} \sum _{x_{i}^{s(c)} \in \bar{\textbf{X}}^{s}} \Phi(\textbf{X}_{i}^{s(c)}) - \frac{1}{n_{t}^{c}} \sum _{x_{j}^{t(c)} \in \bar{\textbf{X}}^{t}} \Phi(\textbf{X}_{j}^{t(c)})\biggl \|_{H}^{2}
\end{equation}

\subsubsection{Low-rank Coding:}
\label{Low-rank coding}
We apply Low-rank coding to the second base model for aligning source and target distributions and decreasing domains discrepancy in Equation (\ref{Equation:1}). We can reach this aim by minimizing the Low-rank formulation, which is shown in Equation (\ref{Equation:5}).
\begin{equation} \label{Equation:5}
   \noindent\footnotesize
   D_{Low\_rank}(\bar{\textbf{X}}^{s},\bar{\textbf{X}}^{t}) =  \parallel \textbf{Z} \parallel_{*} + \lambda \parallel \textbf{E} \parallel_{1}   \;\;\; s.t. \bar{\textbf{X}}^{t} = \bar{\textbf{X}}^{s} \textbf{Z} + \textbf{E},
\end{equation}
where $||.||_*$ is the nuclear norm of a matrix \cite{liu2012robust}. The reconstruction matrix \textbf{Z} and noise matrix \textbf{E} can be optimized by Augmented Lagrange Multiplier (ALM) method \cite{lin2010augmented} through fixing one variable and optimizing the other one until it converges.

\subsubsection{Correlation Alignment (CORAL):}
\label{Correlation alignment}
We apply CORrelation ALignment (CORAL) to the third base model for matching the second-order statistics (covariances) between the data distributions in Equation (\ref{Equation:1}). We can reach this aim by minimizing the CORAL formulation, which is shown in Equation (\ref{Equation:6}).
\begin{equation} \label{Equation:6}
  \noindent\footnotesize
  D_{CORAL}(\bar{\textbf{X}}^{s},\bar{\textbf{X}}^{t}) = \frac{1}{4m^{2}} \biggl \| Cov^{s} - Cov^{t} \biggl \| _{F}^{2},
\end{equation}
where $||.||_{F}$ is the Frobenius norm. m is the dimensions of data. $Cov^{s}$ and $Cov^{t}$ are the covariance matrices for the source and target data, respectively.

\subsection{Meta-learner Model}
As we mentioned before, the goal of stacking models is to learn various base models and combine them via training a meta-learner model to obtain more accurate output predictions based on the multi predictions returned through these base models.
In our classification problem, we choose a MMD based, a Low-rank based, and a Coral based classifiers as base learners, and decide to learn two fully connected layers neural network as a meta-learner model. The meta-learner comprises a fully connected layer of 64 units with ReLU activation and another fully connected layer with softmax activation function as the output layer. The meta-learner model will receive as inputs the outputs of our three base learner models and will learn for returning the final predictions.
So we pursue the following steps:

\textbf{Step 1:} Choose three domain adaptation models as the base learners, and fit them to the training data.

\textbf{Step 2:} For each of these three base learners, make predictions for observations to the validation data.

\textbf{Step 3:} Fit the meta-learner model to the validation data by applying predictions that were made through the base learners as meta-learner inputs.

\textbf{Step 4:} Test the meta-learner model by testing data, and obtain the final predictions.

\section{Experimental Results}
\label{Experimental Results}
We will evaluate our proposed model with retinal fundus images collected from patients with macular degeneration.

\subsection{Implementation Details}
The algorithms were implemented in Python and Pytorch, and all convolutional and pooling layers were fine-tuned based on Pytorch-provided models of ResNet \cite{he2016deep}. The optimization approach was mini-batch stochastic gradient descent (SGD) with momentum of 0.9 and learning rate $\eta_{p} =  \frac{\eta_{0}}{(1+\alpha p)^{\beta}}$ where $p$ was in range [0-1], $\eta_{0} = 0.01$, $\alpha = 10$, and $\beta = 0.75$. The classifiers were trained based on back-propagation with a batch size 32 (minibatch) and the accuracy was obtained at epoch 30.

\subsection{Benchmark Dataset}
We evaluate our model on AREDS benchmark ophthalmic dataset.

\textbf{AREDS} \cite{davis2005age} consists of fundus images from 4757 participants (55-80 years) who represented AMD during follow-up (1992-2005). AREDS dataset contains 14 different classes named, 0: Both-NV-AMD-and-GA, 1: Control, 2: Control-Questionable-1, 3: Control-Questionable-2, 4: Control-Questionable-3, 5: Control-Questionable-4, 6: GA, 7: Large-Drusen, 8: Large-Drusen-Questionable-1, 9: Large-Drusen-Questionable-2, 10: Large-Drusen-Questionable-3, 11: NV-AMD, 12: Other-non-control, 13: Questinable-AMD.

\subsection{Results and Discussions}

\begin{figure}[!t]
\centering
\includegraphics[width=0.7\textwidth]{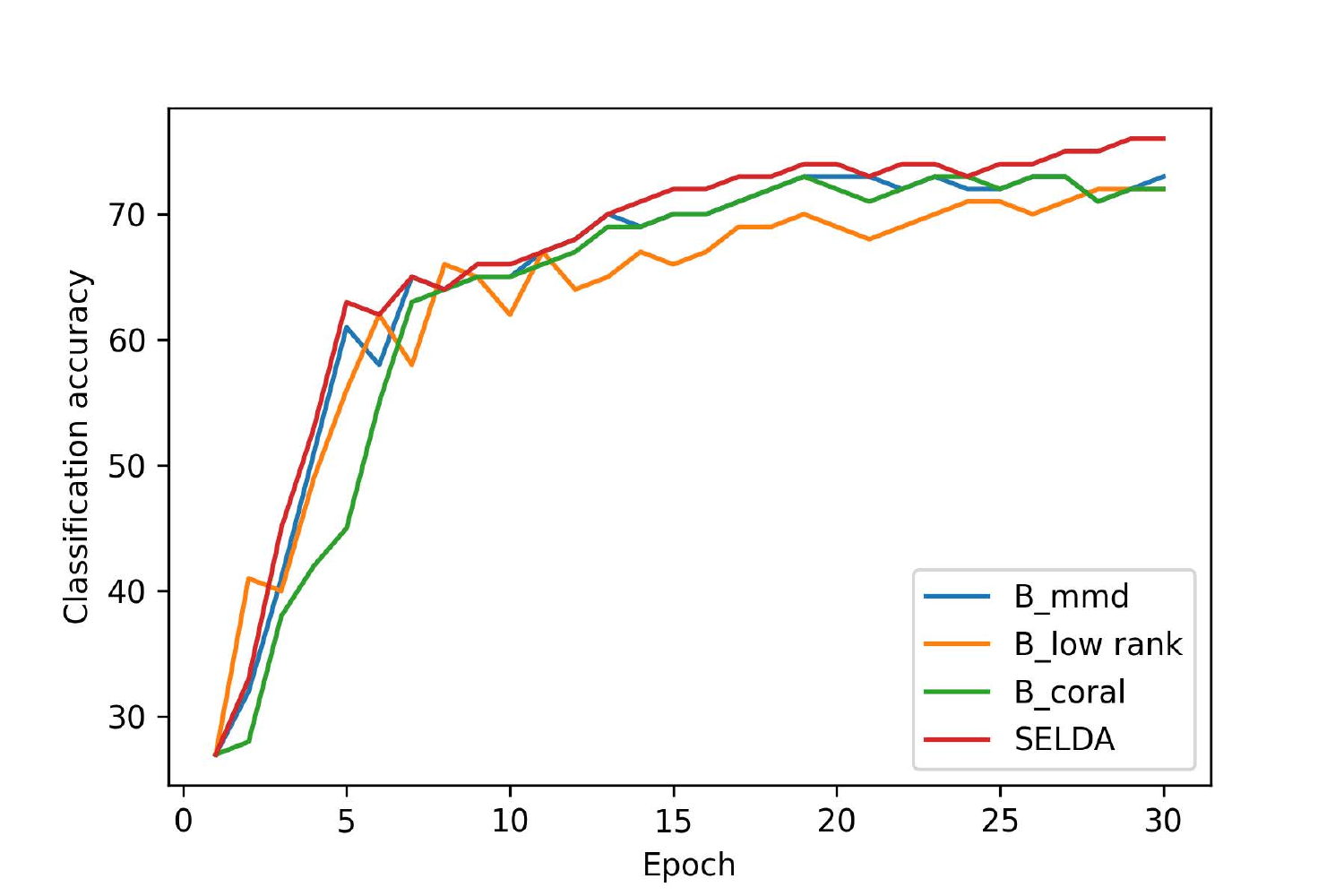}
\caption{Classification accuracy of the base models and ensemble approach versus epoch number based on the AREDS dataset.}
\label{fig:Acc_SELDA}
\end{figure}

\begin{figure}[!t]
\centering
\includegraphics[width=0.95\textwidth]{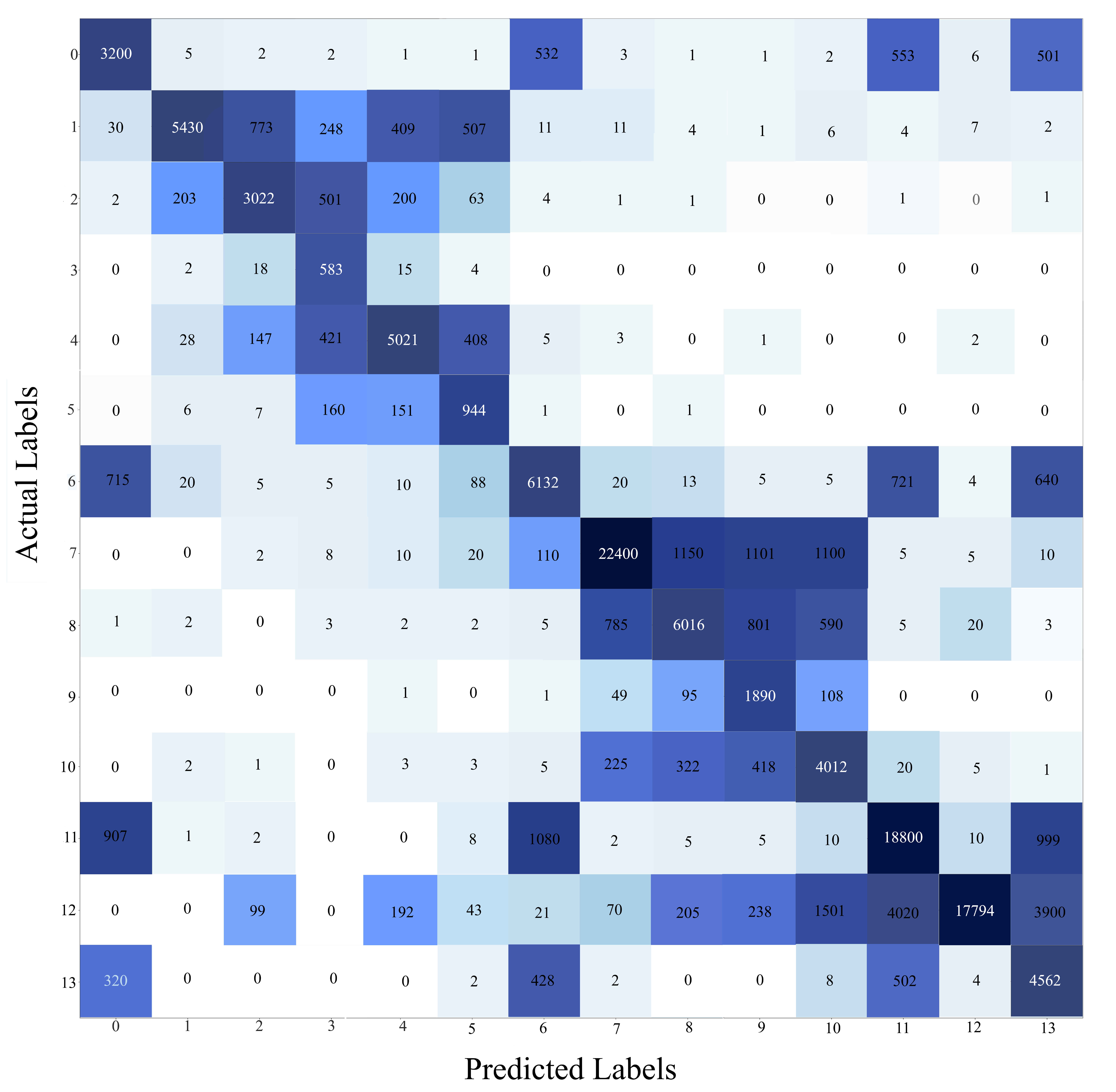}
\caption{The confusion matrix of the proposed model on AREDS dataset.}
\label{fig:AREDS_ConfMat_SELDA}
\end{figure}

We applied our models on AREDS dataset. This dataset includes highly imbalanced classes with substantially greater number of samples in some classes (e.g., large drusen) and significantly small number of samples in other classes. Therefore,we randomly selected 4900 images with an equal number of samples from each class to train our models. We then split the selected images into two parts as AREDS\_ source (80\%) and AREDS\_ target (20\%). However, we tested the model to rest of data to assure generalizability. As some of the eyes included multiple fundus photographs, we assured samples from each eye and patient go to either training, testing, or validation to avoid bias. The classification accuracy of SELDA was obtained 77.85\%. Figure \ref{fig:Acc_SELDA} shows the accuracy versus epoch number and Figure \ref{fig:AREDS_ConfMat_SELDA} shows the confusion matrix of the SELDA. SELDA achieved the highest accuracy compared to the base models and provided an accuracy of 77.85\% for classifying fundus photographs to 14 AMD classes.

Burlina et al. \cite{burlina2017automated} developed several deep learning models to detect four severity levels of AMD based on the AREDS dataset and obtained accuracy ranging from 83.2\% to 91.6\%. However, this model was able to detect only four severity levels while detecting AMD in finer levels has more clinical relevance. In a follow up study, the same team \cite{burlina2018use} developed a deep learning-based model to identify the detailed severity characterization of patients with AMD based on the AREDS dataset and obtained an accuracy level of 59.1\% in identifying 9 different classes. Grassman et al. \cite{grassmann2018deep} developed a framwork based on an ensemble of six different deep learning architectures to identify 9-step (12 classes) grading of ADM based on AREDS dataset and achieved an overall accuracy of 63.3\%. However, our model was able to identify 14 different classes of AMD and achieved an accuracy of 77.85\%.

Peng et al. \cite{peng2019deepseenet} developed a deep learning model to detect different severity levels of AMD based on same AREDS dataset and obtained an accuracy of 67.1\% while the accuracy of SELDA was about 10\% higher than their model too.

Results indicate that SELDA outperforms state-of-the-art \cite{peng2019deepseenet,burlina2017automated}. This is achieved by iteratively reducing the domain discrepancy and effectively propagating the class labels. This could be justified by the fact that SELDA inherits the capabilities of each of the base learner methods; MMD guarantees to minimize marginal and conditional distributions difference between the source and target domains, the low-rank representation extracts more relevant information shared between domains by constructing the block-wise structure,and CORAL tries to align the covariances of the source and target domains to mitigate domain discrepancy.

The ablation study was performed to evaluate the efficiency of the proposed method.
First, we run our model only by using $B_{MMD}$, second, only by using $B_{Low_rank}$, third, only by using $B_{Coral}$, and finally, by ensembling on all three base learners. The accuracies of $B_{MMD}$, $B_{Low_rank}$, $B_{Coral}$, and SELDA were obtained 73.33\%, 72.61\%, 72.35\%, and 77.85\%, respectively. As it is seen, the best results were obtained using SELDA, which utilizes all three base learner models.
The results show that our proposed model has learned to extract important features from the macular region of the fundus images. Furthermore, because the network has learned the features which were most predictive for the related class, it is feasible that the model is utilizing features previously to be unknown or have been ignored by humans which may be highly predictive of certain AREDS classes, so it can be efficiently trained to detect specific disease-related changes on fundus images.

\section{Conclusions}
\label{Conclusions}
In this paper, we rethink domain adaptation problem and propose stacking ensemble learning by utilizing MMD-based CNN, low rank-based CNN, and CORAL-based CNN base DA learners and a meta-learner to address domain shift challenge in ophthalmology and diagnosis of eye diseases. We utilize a two-fully connected layer network as a meta-learner model to stack the output predictions of these three well-performing DA models to obtain high accuracy in ophthalmic image classification tasks. The proposed model jointly inherits the capabilities of each of the base learner models, efficiently. Extensive experimental results and analyses on AREDS visual benchmark dataset have illustrated the effectiveness of our model.

%
%
\bibliographystyle{spphys}
\bibliography{bibsource}

\end{document}